\documentclass[nohyperref]{article}

\usepackage{microtype}
\usepackage{graphicx}
\usepackage{subfigure}
\usepackage{booktabs} %
\usepackage{units}
\usepackage{placeins}
\usepackage[textsize=tiny]{todonotes}

\usepackage{hyperref}

\usepackage[accepted]{icml2025}

\usepackage{amsmath}
\usepackage{amssymb}
\usepackage{mathtools}
\usepackage{amsthm}
\usepackage{arydshln}

\usepackage[capitalize,noabbrev]{cleveref}

\theoremstyle{plain}

\theoremstyle{definition}

\theoremstyle{remark}

\usepackage{amsmath,amsfonts,bm}

\def\eqref#1{equation~\ref{#1}}

\def\1{\bm{1}}

\DeclareMathAlphabet{\mathsfit}{\encodingdefault}{\sfdefault}{m}{sl}
\SetMathAlphabet{\mathsfit}{bold}{\encodingdefault}{\sfdefault}{bx}{n}

\usepackage{xparse}
\usepackage{xspace}
\usepackage{algorithm}%
\usepackage{fixltx2e}
\usepackage{amsmath}
\usepackage{tikz}
\usepackage{float}
\usepackage{multirow}
\usepackage{multicol}
\usepackage{colortbl}

\usepackage{microtype}
\usepackage{graphicx}
\usepackage{subfigure}
\usepackage{booktabs} %
\usepackage[shortlabels]{enumitem}

\usepackage{hyperref,amssymb,enumitem}

\usepackage{wrapfig}

\setlist[itemize]{leftmargin=*}
\setlist[enumerate]{leftmargin=*}

\usepackage{amsmath}

\graphicspath{{images/}}

\newif\ifdraft
\draftfalse

\usepackage{fancyvrb}

\RecustomVerbatimCommand{\VerbatimInput}{VerbatimInput}%
{fontsize=\footnotesize,
 frame=lines,  %
 framesep=2em, %
 commandchars=\|\(\), %
 commentchar=*        %
}

\makeatletter
\def\adl@drawiv#1#2#3{%
        \hskip.5\tabcolsep
        \xleaders#3{#2.5\@tempdimb #1{1}#2.5\@tempdimb}%
                #2\z@ plus1fil minus1fil\relax
        \hskip.5\tabcolsep}
\newcommand{\cdashlinelr}[1]{%
  \noalign{\vskip\aboverulesep
           \global\let\@dashdrawstore\adl@draw
           \global\let\adl@draw\adl@drawiv}
  \cdashline{#1}
  \noalign{\global\let\adl@draw\@dashdrawstore
           \vskip\belowrulesep}}
\makeatother

\ifdraft
\newcommand{\mynote}[1]{\textcolor{red}{[note: #1]}}

\newcommand{\mytodo}[1]{\textcolor{red}{[todo: #1]}}
\newcommand{\mycomment}[1]{\textcolor{red}{[comment: #1]}}
\newcommand{\chris}[1]{\textcolor{red}{Chris: #1}}
\newcommand{\ahmad}[1]{\textcolor{blue}{Ahmad: #1}}
\newcommand{\vinith}[1]{\textcolor{blue}{Vinith: #1}}
\newcommand{\yunxiang}[1]{\textcolor{cyan}{Yunxiang: #1}}
\newcommand{\xiao}[1]{\textcolor{blue}{xiao: #1}}
\definecolor{chocolate(traditional)}{rgb}{0.48, 0.25, 0.0}

\definecolor{darkpastelgreen}{rgb}{0.01, 0.75, 0.24}
\newcommand{\natalie}[1]{\textcolor{darkpastelgreen}{natalie: #1}}
\definecolor{pistachio}{rgb}{0.58, 0.77, 0.45}

\newcommand{\jonas}[1]{\textcolor{magenta}{[Jonas: #1]}}

\newcommand{\adelin}[1]{\textcolor{red}{[Adelin: #1]}}
\newcommand{\mohammad}[1]{\textcolor{red}{[Mohammad: #1]}}

\definecolor{amber(sae/ece)}{rgb}{1.0, 0.49, 0.0}
\newcommand\adam[1]{{\textcolor{red}{[Adam: #1]}}}
\newcommand{\sierra}[1]{\textcolor{blue}{[Sierra: #1]}}
\newcommand{\armin}[1]{\textcolor{cyan}{[Armin: #1]}}
\newcommand{\nikita}[1]{\textcolor{cyan}{[Nikita: #1]}}

\else

\newcommand{\chris}[1]{}
\newcommand{\vinith}[1]{}
\newcommand{\adam}[1]{}
\newcommand{\yunxiang}[1]{}
\newcommand{\natalie}[1]{}
\newcommand{\jonas}[1]{}
\newcommand{\adelin}[1]{}
\newcommand{\mynote}[1]{}
\newcommand{\xiao}[1]{}
\newcommand{\mytodo}[1]{}
\newcommand{\mycomment}[1]{}
\newcommand{\ahmad}[1]{}
\newcommand{\mohammad}[1]{}
\newcommand{\sierra}[1]{}
\newcommand{\armin}[1]{}
\newcommand{\nikita}[1]{}
\fi

\usepackage{tikz}
\usetikzlibrary{mindmap}
\usepackage[linguistics]{forest}

\newcommand{\mytitle}{Localizing and Mitigating Memorization in Image Autoregressive Models}
\icmltitlerunning{\mytitle}

\begin{document}

\twocolumn[
\icmltitle{\mytitle}

\begin{icmlauthorlist}
\icmlauthor{Aditya Kasliwal}{comp}
\icmlauthor{Franziska Boenisch}{comp}
\icmlauthor{Adam Dziedzic}{comp}
\end{icmlauthorlist}

\icmlaffiliation{comp}{CISPA Helmholtz Center for Information Security}

\icmlcorrespondingauthor{Aditya Kasliwal}{kasliwal.aditya@cispa.de}
\icmlcorrespondingauthor{Franziska Boenisch}{boenisch@cispa.de}
\icmlcorrespondingauthor{Adam Dziedzic}{adam.dziedzic@cispa.de}

\icmlkeywords{Machine Learning, ICML, Memorization}

\vskip 0.3in
]

\printAffiliationsAndNotice{} %

\begin{abstract}

Image AutoRegressive (IAR) models have achieved state-of-the-art performance in speed and quality of generated images. However, they also raise concerns about memorization of their training data and its implications for privacy. This work explores where and how such memorization occurs within different image autoregressive architectures by measuring a fine-grained memorization. The analysis reveals that memorization patterns differ across various architectures of IARs. In hierarchical per-resolution architectures, it tends to emerge early and deepen with resolutions, while in IARs with standard autoregressive per token prediction, it concentrates in later processing stages. These localization of memorization patterns are further connected to IARs’ ability to memorize and leak training data. By intervening on their most memorizing components, we significantly reduce the capacity for data extraction from IARs with minimal impact on the quality of generated images. These findings offer new insights into the internal behavior of image generative models and point toward practical strategies for mitigating privacy risks.
\end{abstract}

\section{Introduction}
Image Autoregressive (IAR) models, like Visual Autoregressive Modeling (VAR) \citep{tian2024visual} and Randomized Autoregressive (RAR) models \citep{yu2024randomized}, are the state of the art in generative modeling, with superior image quality and generation efficiency compared to other frameworks~\citep{han2024infinity}. Yet, this success is tempered by their tendency to memorize training data \citep{kowalczuk2025privacy}, which engenders privacy concerns such as possible data leakage and facilitating extraction attacks. For generative models, which aim for novelty, such memorization is a fundamental issue \citep{chavhan2024memorized}.

Understanding \textit{where} memorization occurs in complex networks is crucial for mitigation \citep{maini2023can, wanglocalizing, wang2024memorization, hintersdorf2024finding, wang2025captured}. This paper studies memorization localization in VAR (VAR-d16, VAR-d30) and RAR (Base, XXL) models using the UnitMem metric \citep{wanglocalizing}, which quantifies unit-level memorization efficiently without labels. For VAR, we examine memorization per block across generation scales; for RAR, analysis is block-wise due to its token-sequential nature.

Our findings reveal that VAR's memorization shifts from initial blocks at coarse scales to deeper blocks at finer scales. RAR models show memorization concentrated in middle and later blocks. We validate these localization patterns by scaling down the weights of a targeted percentage of high-UnitMem neurons by half. This intervention significantly reduces extractable images (e.g., for VAR-d30, from 672 to 110, and for RAR-XXL, from 75 to 26) with a controlled impact on FID, confirming UnitMem's accuracy in pinpointing memorization-critical neurons. This work offers new insights into IAR memorization dynamics and validates a method for their analysis.

\section{Background}

\textbf{Image Autoregressive (IAR) Models.} IARs generate images sequentially, typically token by token, building upon foundational concepts in autoregressive modeling for generative tasks \citep{van2016pixel}.
\textbf{VAR} \citep{tian2024visual} employs a next-scale prediction strategy, processing images hierarchically from coarse to fine resolutions. 
\textbf{RAR} \citep{yu2024randomized} often incorporates bidirectional context, drawing from masked language modeling, by using permuted token sequences and bidirectional attention.

\textbf{Memorization in Generative Models.} It refers to a model's ability to reproduce or reconstruct training instances with high fidelity \citep{song2017machine,wei2024memorization}. This phenomenon poses significant privacy risks, as memorized data, if sensitive, could be inadvertently leaked or maliciously extracted \citep{kowalczuk2025privacy, carlini2021extracting}. While extensively studied in diffusion models (DMs) \citep{carlini2023extracting, somepalli2023understanding}, IARs have also been shown to exhibit significant, sometimes even more pronounced, memorization tendencies \citep{kowalczuk2025privacy}.

\textbf{Localizing Memorization.} This process seeks to identify network components responsible for storing specific training instances \citep{maini2023can, hintersdorf2024finding, stoehr2024localizing}. Central to our work is the \textbf{UnitMem} metric \citep{wanglocalizing}, which quantifies unit-level memorization by measuring activation sensitivity to specific inputs. UnitMem is highly scalable for large models as it requires only a single forward pass and no labels. \citet{wanglocalizing} previously used it to show that memorizing units in vision encoders can be distributed and vary with layer depth.

\textbf{Data Extraction as Validation.} The ability to extract training data is a direct consequence of memorization. \citet{kowalczuk2025privacy} demonstrate significant data extraction from IARs. Modifying neurons identified as highly memorizing should thus impede extraction, serving as a functional validation for localization techniques.

\section{Method}
\label{sec:method}
 Our methodology for localizing memorization in Image Autoregressive (IAR) models involves applying the UnitMem metric \citep{wanglocalizing} to VAR and RAR architectures and subsequently validating these localization results through data extraction experiments following~\citet{kowalczuk2025privacy}.

 \subsection{Models and Datasets}

We focus on two prominent IAR architectures: Visual Autoregressive Modeling (VAR), using two variants, VAR-d16 and VAR-d30, which represent its smallest and largest configurations respectively \citep{tian2024visual}; and Randomized Autoregressive (RAR) models, analyzing both RAR-Base and RAR-XXL, which are similarly the smallest and largest available variants \citep{yu2024randomized}. All models are trained on the large-scale ImageNet-1k dataset \cite{russakovsky2015imagenet}. For UnitMem calculation, we utilize a 1\% subset of the ImageNet-1k training data, sampled evenly from all 1000 classes. This subsetting strategy is adopted to reduce computational overhead; experiments with 1\%, 5\%, 10\%, and 20\% subsets yield similar memorization patterns, validating the use of the 1\% subset for our analysis as sufficient (see Appendix~\ref{app:experiments} for more details).

 \subsection{UnitMem Calculation for IARs}
 The UnitMem metric \citep{wanglocalizing} quantifies the memorization of an individual unit (neuron or channel) by measuring its sensitivity to specific training data points. It is defined as:
$$ \text{UnitMem}_{\mathcal{D}'}(u) = \frac{\mu_{max,u} - \mu_{-max,u}}{\mu_{max,u} + \mu_{-max,u}},$$
where $\mu_{max,u}$ is the maximum activation of unit $u$ for a specific data point $x_k$ in a training subset $\mathcal{D}'$, and $\mu_{-max,u}$ is the mean activation of $u$ for all other data points in $\mathcal{D}' \setminus \{x_k\}$. Following the methodology of \citet{wanglocalizing}, the activation for any given data point $x_i$ is computed by averaging the neuron's activations over 10 forward passes of $x_i$, each with a different data augmentation identical to those used during model training.
 
 Both VAR and RAR model blocks typically consist of an attention layer followed by two fully connected layers (fc1 and fc2). Our UnitMem analysis focuses on the neurons within the fc1 layer of each block. This is because fc1 employs a GELU activation function, whereas fc2 has no activation function. The original UnitMem metric is designed to work with ReLU activations; to adapt it for GELU, which can produce negative values, we use the absolute value of the activations when computing $\mu_{max,u}$ and $\mu_{-max,u}$. This ensures that the magnitude of activation, regardless of sign, contributes to the memorization score.

 To adapt UnitMem for IARs, which involve iterative application of the same model components (blocks) for generating different parts of an image (tokens or scales), we employ a \textit{teacher-forced inference} approach. Similar to teacher-forced training, during the calculation of activations for UnitMem, the model receives ground-truth inputs from the preceding step rather than its own potentially erroneous predictions. This ensures that we are measuring the model's inherent activation patterns and memorization tendencies with respect to the actual training data distribution at each stage of generation.

 \subsubsection{UnitMem for VAR Models}
 VAR models generate images hierarchically across multiple scales (10 scales). The same set of transformer blocks is reused to predict tokens for each subsequent, finer scale, conditioned on the tokens from the immediately preceding coarser scale.

 To handle the multiple activations a single neuron might exhibit due to this iterative process, we calculate UnitMem in a scale-wise manner for VAR. For each of the $S$ scales:
 \begin{enumerate}
 \item \textbf{Input Provisioning}: For generating tokens at a given scale $s$, the model receives the ground-truth tokens from the previous scale $s-1$.
 \item \textbf{Activation Collection}: As the model processes the input to predict all tokens within the current scale $s$, we record the activations of each neuron in the fc1 layer of every block. Since all tokens within a scale $s$ are predicted based on the same input from scale $s-1$ (differing only by their positional encodings), we obtain multiple activation values for each neuron corresponding to each token being generated within that scale.
 \item \textbf{Scale-Specific Mean Activation}: For each training image $x$ and each neuron $u$ in fc1, we compute its mean absolute activation $\bar{a}_{u,s}(x)$ at scale $s$ by averaging its absolute activations across all tokens generated within that scale $s$.
 \item \textbf{UnitMem per Scale}: Using these scale-specific mean absolute activations $\bar{a}_{u,s}(x)$, we then compute $\text{UnitMem}_s(u)$ for each neuron $u$ at each scale $s$ following the standard UnitMem formula, using a subset of training images $\mathcal{D}'$.
 \end{enumerate}
 This process is detailed in Algorithm~\ref{alg:var_unitmem_state_for} (see Appendix~\ref{appendix:algorithm}). The final memorization score for a neuron can then be analyzed per scale.

 \subsubsection{UnitMem for RAR Models}
 RAR models generate images by predicting tokens sequentially, where the input for predicting token $t_i$ consists of all previously generated (or ground-truth, in our teacher-forced setting) tokens $t_1, \dots, t_{i-1}$.

 Therefore, for RAR models, we calculate UnitMem based on the model's state when it predicts the \textit{final token} of a sequence (e.g., the 256th token if images are represented by 256 tokens).
 \begin{enumerate}
 \item \textbf{Input Provisioning}: For each training image $x$, we provide all its ground-truth tokens $t_1, \dots, t_{N-1}$ as input to the model.
 \item \textbf{Activation Collection for Last Token}: We then record the activations of each neuron in the fc1 layer of every block as the model processes this context ($t_1, \dots, t_{N-1}$) to predict the final token $t_N$. Let this activation for neuron $u$ and image $x$ be $a_{u,N}(x)$.
 \item \textbf{UnitMem Calculation}: Using the absolute value of activations $|a_{u,N}(x)|$ (which represent the mean absolute activation for the last token prediction step, as it is a single step per image), we compute $\text{UnitMem}(u)$ for each neuron $u$ in fc1 following the standard UnitMem formula over the training subset $\mathcal{D}'$.
 \end{enumerate}
 This approach captures the memorization patterns of neurons when the model has seen almost the entire context of an image.

 \subsection{Validation via Data Extraction}
 To verify whether the localization of memorization by UnitMem is accurate, we assess the impact of modifying high-UnitMem neurons on the model's ability to reproduce training samples. We follow the data extraction methodology outlined by~\citet{kowalczuk2025privacy}:
 \begin{enumerate}
 \item \textbf{Candidate Identification}: Promising training samples that are likely memorized are identified.
 \item \textbf{Prefix-based Generation}: For each candidate sample, a prefix of its tokenized representation is fed to the IAR. The model then autoregressively generates the remaining tokens to complete the image.
 \item \textbf{Similarity Assessment}: The generated image is compared to the original candidate image using SSCD \citep{pizzi2022self}. If similarity $>$ 0.75, it is considered extracted.
 \end{enumerate}
 We perform this data extraction procedure under two conditions: first, using the \textbf{Original Models} (unmodified VAR and RAR models) to establish a baseline number of extractable images, and second, using \textbf{Modified Models}. For the modified condition, after identifying the top 10\% of neurons in fc1 layers with the highest UnitMem scores (aggregated across scales for VAR, and from the last token prediction for RAR), we intervene on these neurons. Having explored several modification strategies (including zeroing out weights and scaling weights/biases by different factors, with further details in Appendix~\ref{app:experiments}), we found that halving the weights (scale by 0.5) of these specific neurons without altering their biases was the most effective. This approach best balanced extraction reduction with minimal FID impact, unlike other strategies that led to more substantial FID degradation or less effective extraction reduction. The data extraction attack is then re-run on these modified models.
 
 Next, in the empirical evaluation, we show a significant reduction in the number of extractable images from the modified models with minimal impact on overall generation quality (measured by FID), which validates that UnitMem correctly identifies neurons crucial for memorizing training instances.

 \section{Experiments and Results}
 \label{sec:empirical_evaluation} %

 \subsection{Experimental Setup}
 We apply the UnitMem calculation methodology to VAR-d16, VAR-d30, RAR-Base, and RAR-XXL models. The 1\% subset of the ImageNet training dataset, evenly sampled across classes, is used for UnitMem computation. All activations are collected from fc1 layers using the teacher-forced inference approach, and absolute activation values are used.

 \subsection{Memorization Localization Patterns}
 The distribution of memorization, as quantified by UnitMem scores, is analyzed for each model. For visualization, we generate heatmaps representing the sum of UnitMem scores for all neurons within the fc1 layer of each block of the models.

 \textbf{VAR Models (VAR-d16 and VAR-d30)}:
 The heatmaps for VAR models (\Cref{fig:var_heatmap_placeholder} for VAR-d16 and \Cref{fig:var_heatmap_appendix_d30} in \Cref{appendix:additional_plots} for VAR-d30) illustrate the sum of UnitMem scores per block, further aggregated or visualized across the 10 generation scales.
 A key observation for both VAR-d16 and VAR-d30 is a clear trend related to the generation scale. At the initial, coarser scales, memorization is predominantly concentrated in the earlier blocks. As generation progresses to finer scales, the locus of high memorization shifts towards deeper blocks.

 \begin{figure} %
 \centering
 \includegraphics[width=0.5\textwidth]{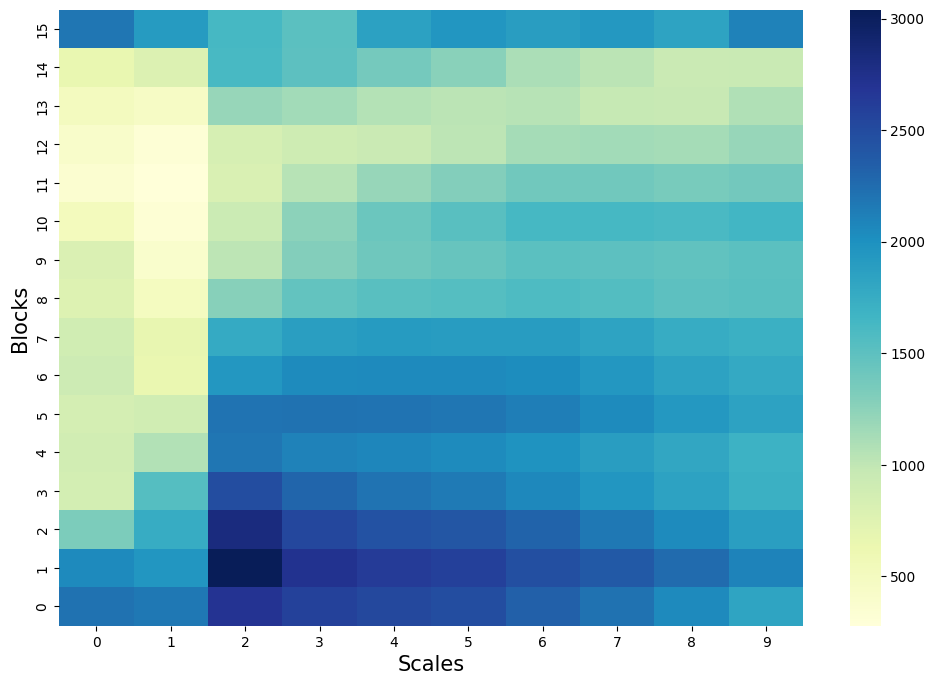}
 \caption{Heatmap illustrating the sum of UnitMem scores per block (fc1 layer) across different scales for VAR-d16 model. Darker colors indicate higher memorization.}
 \label{fig:var_heatmap_placeholder}
 \end{figure}

 \textbf{RAR Models (RAR-Base and RAR-XXL)}:
 For RAR models, where UnitMem is calculated based on the prediction of the final token, the heatmaps (\Cref{fig:rar_heatmap_placeholder} for RAR-Base and \Cref{fig:rar_heatmap_appendix_xx} in \Cref{appendix:additional_plots} for RAR-XXL) show the sum of UnitMem scores per block (fc1 layer).
 In both RAR-Base and RAR-XXL, higher memorization scores tend to concentrate in the middle and later blocks.

 \begin{figure} %
 \centering
 \includegraphics[width=0.5\textwidth]{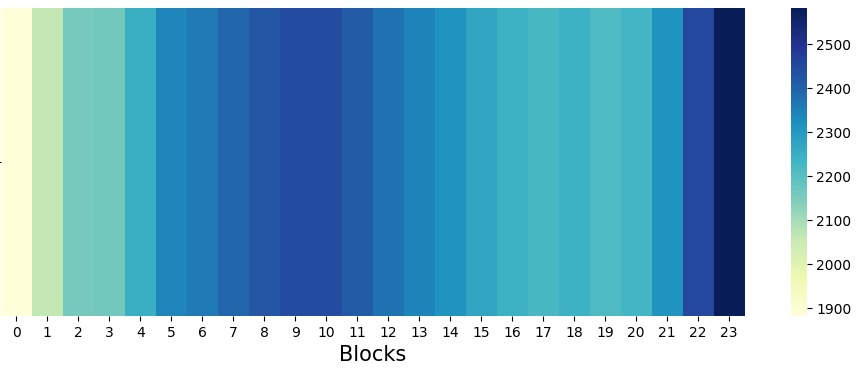}
 \caption{Heatmap illustrating the sum of UnitMem scores per block (fc1 layer) for RAR-Base model, based on last token prediction. Darker colors indicate higher memorization.}
 \label{fig:rar_heatmap_placeholder}
 \end{figure}

 \subsection{Validation through Data Extraction}
 To validate the localization results, we performed data extraction attacks.

 \textbf{Baseline Extraction}:
 Replicating \citet{kowalczuk2025privacy}, we extract \textbf{672} images from VAR-d30 and \textbf{75} from RAR-XXL.

\textbf{Extraction after Neuron Modification}:
To validate the localization, we intervened on fc1 neurons identified with the highest UnitMem scores by halving their weights. Our analysis focuses on the impact of modifying the top 10\% of such neurons in VAR-d30 and the top 5\% in RAR-XXL (a comprehensive analysis of varying intervention strengths for RAR is provided in Appendix~\ref{app:experiments}).

This intervention substantially reduced extractable images, confirming UnitMem’s ability to identify critical memorizing components. For VAR-d30, extractable images were cut by 83.6\% (from 672 to 110) with a marginal FID increase (1.97$\rightarrow$2.58). For RAR-XXL, extractions were reduced by 65.3\% (from 75 to 26), though with a more notable FID increase from 1.48 to 5.12. These results demonstrate effective mitigation of data extraction. However, they also highlight a trade-off with generative quality, particularly for RAR-XXL (this trade-off is further explored for RAR models in Appendix~\ref{app:experiments}).

\section{Conclusions}

In this work, we successfully applied the UnitMem metric to localize memorization in VAR and RAR image autoregressive models. Our findings reveal distinct, architecture-dependent memorization patterns: VAR models exhibit a scale-dependent shift in memorization from early to deeper blocks, while RAR models concentrate memorization in their middle to later stages. Crucially, modifying the neurons UnitMem identified as most memorizing significantly reduced extractable training images with minimal impact on generation quality, thereby validating UnitMem as an effective tool for pinpointing memorization-critical components in IARs. These insights pave the way for more targeted strategies to mitigate privacy risks and deepen understanding of IAR internal dynamics, while also guiding future work such as leveraging these localized findings for efficient model pruning or novel regularization techniques targeting memorizing neurons.

\bibliography{main}
\bibliographystyle{icml2025}

\newpage
\appendix
\twocolumn %
 \section{Details on the Experimental Setup}
 \label{app:experiments}

 \textbf{UnitMem Computation Subset}:
 As mentioned in Section~\ref{sec:method}, we used a 1\% subset of the ImageNet-1k dataset for UnitMem calculation. This subset was created by randomly sampling an equal number of images from each of the 1000 classes to ensure balanced representation. The primary reason for using a subset was to significantly reduce the computational time required for the extensive forward passes needed for UnitMem. We conducted preliminary experiments comparing the localization patterns derived from 1\%, 5\%, 10\%, and 20\% subsets of ImageNet-1k. The resulting heatmaps and distributions of high-UnitMem neurons were qualitatively and quantitatively similar across these subset sizes. Given this consistency, we proceeded with the 1\% subset for all reported results, confident that it provides a reliable estimate of memorization localization while being computationally feasible.

\textbf{A Note on Comparability of Extracted Samples.}
Regarding our baseline data extraction results (e.g., 672 images from VAR-d30 and 75 from RAR-XXL, as reported in Section~\ref{sec:empirical_evaluation}), while we closely follow the methodology and utilize the same codebase as \citet{kowalczuk2025privacy}, the precise set of individual training samples extracted may not be an exact one-to-one match with those potentially identified in their original work. The experiments conducted by \citet{kowalczuk2025privacy} involved a distributed computing setup, which, led to the exclusion of certain samples. Our replication was performed in a non-distributed environment. Such differences in large-scale experimental execution and data handling can result in variations in the specific instances identified as extracted, even when overall quantitative findings and the efficacy of the extraction methodology are comparable. This context is important when considering fine-grained comparisons of the exact data instances.

 \textbf{Neuron Intervention Strategy}:
 In our validation experiments (Section~\ref{sec:empirical_evaluation}), we intervened on neurons in fc1 layers identified by UnitMem as highly memorizing. Several intervention strategies were explored to find an optimal balance between reducing data extraction and preserving model utility:
\begin{itemize}
\item \textbf{Zeroing out weights}: Setting the weights of identified neurons to zero. This did not yield significant benefits in reducing extractable images, possibly due to the models' inherent robustness from dropout regularization used during training, or it may have overly damaged model utility.
\item \textbf{Scaling weights and biases}: We experimented with scaling both weights and biases of the identified neurons by various factors (e.g., 0.1, 0.25, 0.5, 0.75).
\end{itemize}
The most effective general strategy, in terms of maximizing the reduction in extractable images while minimizing the impact on FID, was scaling down the weights of the identified fc1 neurons by a factor of 0.5 (halving them), without altering their biases. Scaling down biases alongside weights resulted in a more substantial, undesirable increase in FID, indicating a greater impact on the model's general generative capabilities. Halving only the weights of high-UnitMem fc1 neurons provided the best trade-off among these explored methods.

For VAR-d30, this weight-halving strategy was applied to the top 10\% of high-UnitMem neurons in fc1 layers. As reported in the main paper (Section~\ref{sec:empirical_evaluation}), this effectively reduced extractable images from 672 to 110 with a marginal FID increase from 1.97 to 2.58.

For RAR-XXL, we conducted a more detailed analysis to investigate the trade-off between the extent of intervention (i.e., the percentage of top memorizing neurons modified) and its impact on both data extraction and model utility. The original RAR-XXL model (FID: 1.48) allowed for the extraction of 75 images. We then applied the same weight-halving strategy to different proportions of the top fc1 neurons identified by UnitMem:
\begin{itemize}
\item Scaling the weights of the top 10\% of neurons: Extractable images dropped from 75 to 13 (an 82.7\% reduction). However, the FID increased significantly from 1.48 to 7.3.
\item Scaling the weights of the top 5\% of neurons: Extractable images dropped from 75 to 26 (a 65.3\% reduction), with the FID increasing from 1.48 to 5.12. This level of intervention was selected for detailed discussion in the main paper (Section~\ref{sec:empirical_evaluation}) as it offered a substantial reduction in extractable data while the FID impact, though notable, was less severe than the 10\% intervention.
\item Scaling the weights of the top 1\% of neurons: Extractable images dropped from 75 to 68 (a 9.3\% reduction), with the FID increasing from 1.48 to 3.21.
\end{itemize}
These graduated results for RAR-XXL clearly demonstrate a trade-off: more aggressive intervention on memorizing neurons leads to greater reductions in extractable data but also more significantly impacts the model's overall generation quality as measured by FID. This suggests that while highly memorizing neurons are indeed critical for storing specific training instances, they may also contribute to the model's broader generative capabilities. This observation aligns with broader discussions where it is posited that some degree of memorization might be intertwined with a model's ability to generalize or achieve high performance. Our findings suggest that simply eliminating all memorizing components might not always be optimal if it unduly harms model utility, pointing towards a need for carefully calibrated interventions based on the specific model and privacy-utility requirements.

 \textbf{Target Layer for UnitMem and Intervention}:
 Both VAR and RAR architectures have blocks typically structured as: an attention mechanism, followed by a first fully connected layer (fc1), and then a second fully connected layer (fc2). Our UnitMem calculations and subsequent interventions were specifically targeted at the neurons within the fc1 layer. The fc1 layer in these models uses a GELU activation function. In contrast, the fc2 layer lacks an activation function, meaning its output is not bounded or transformed non-linearly in the same way. The UnitMem metric, originally developed with ReLU activations in mind (which are non-negative), relies on activation magnitudes. To adapt UnitMem for GELU, which can output negative values, we used the absolute value of the neuron's activation when calculating $\mu_{max,u}$ and $\mu_{-max,u}$. This captures the strength of a neuron's response irrespective of its sign, which is appropriate for measuring sensitivity. Applying UnitMem to fc2 layers with no activation was deemed less suitable as the unbounded outputs might not align well with the metric's underlying assumptions about activation distributions.

 \section{Algorithm for UnitMem Calculation in VAR}
 \label{appendix:algorithm}
 Algorithm~\ref{alg:var_unitmem_state_for} details the per-scale UnitMem calculation for VAR models.

 \begin{algorithm}[H]
 \caption{Per-Scale UnitMem Calculation for VAR}
 \label{alg:var_unitmem_state_for}
 \begin{algorithmic}[1] %
  \STATE Input: VAR model $M$, training subset $\mathcal{D}'$, number of scales $S$.
  \FOR{each training image $x \in \mathcal{D}'$}
    \FOR{each scale $s = 1 \dots S$}
      
          \STATE Provide ground-truth tokens from scale $s-1$ as input to $M$.
          \STATE Collect absolute activations $A_{u,s,t}(x)$ for each neuron $u$ in fc1 layers for all tokens $t$ at scale $s$.
          \STATE Compute mean absolute activation $\bar{a}_{u,s}(x) = \text{mean}_t(A_{u,s,t}(x))$ for each neuron $u$.
      \ENDFOR
      \ENDFOR
    \FOR{each scale $s = 1 \dots S$}
      \FOR{each neuron $u$ in fc1 layers}
          \STATE Find $x_k = \arg\max_{x \in \mathcal{D}'} \bar{a}_{u,s}(x)$.
          \STATE $\mu_{max,u,s} = \bar{a}_{u,s}(x_k)$.
          \STATE $\mu_{-max,u,s} = \text{mean}_{x \in \mathcal{D}' \setminus \{x_k\}} \bar{a}_{u,s}(x)$.
          \STATE $\text{UnitMem}_s(u,s) = (\mu_{max,u,s} - \mu_{-max,u,s}) / (\mu_{max,u,s} + \mu_{-max,u,s})$.
      \ENDFOR
    \ENDFOR
 \end{algorithmic}
 \end{algorithm}

 \section{Additional Memorization Localization Heatmaps}
\label{appendix:additional_plots}
This section contains supplementary heatmaps for VAR-d30 and RAR-XXL models, corresponding to the analysis presented in Section~\ref{sec:empirical_evaluation}. Notably, for the VAR-d30 model at scale 0 (the initial generation stage), memorization appears relatively low and is concentrated in the very early blocks. This pattern is attributed to the nature of the input at this stage: scale 0 generation is conditioned solely on a class token, providing limited specific instance information for the model to memorize or process deeply.

 \begin{figure}[H]
 \centering
 \includegraphics[width=0.5\textwidth]{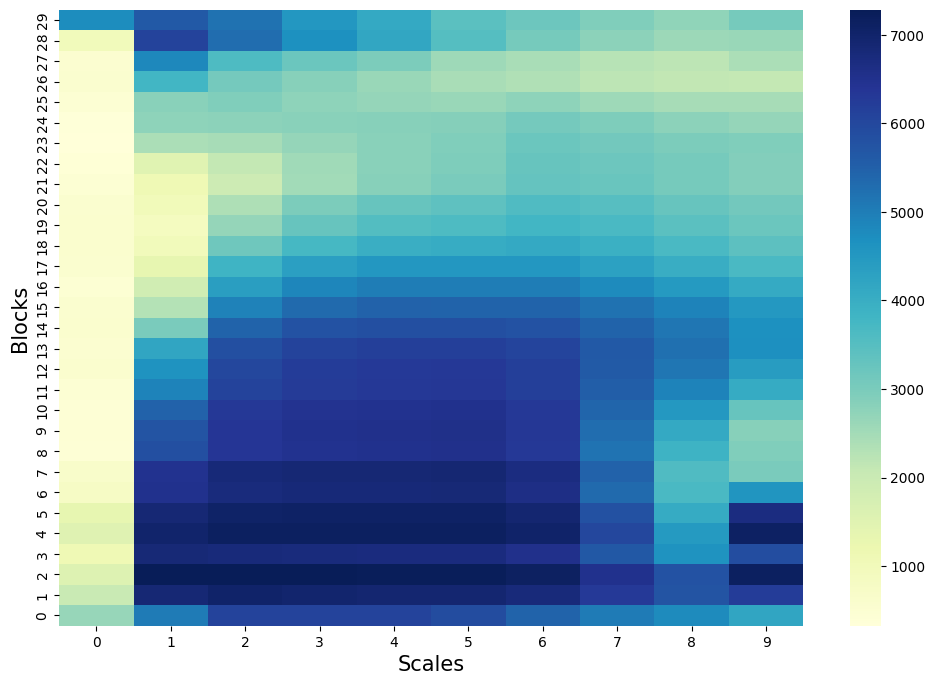} %
 \caption{Heatmap illustrating the sum of UnitMem scores per block (fc1 layer) across different scales for VAR-d30. Darker colors indicate higher memorization.}
 \label{fig:var_heatmap_appendix_d30}
 \end{figure}

 \begin{figure}[H]
 \centering
 \includegraphics[width=0.5\textwidth]{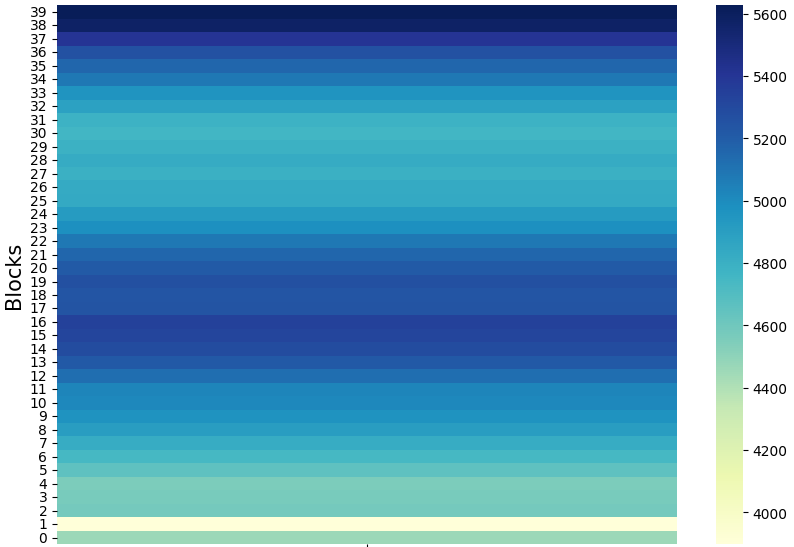} %
 \caption{Heatmap illustrating the sum of UnitMem scores per block (fc1 layer) for RAR-XXL, based on last token prediction. Darker colors indicate higher memorization.}
 \label{fig:rar_heatmap_appendix_xx}
 \end{figure}

 \section{Neuron-wise Memorization Analysis}
 \label{appendix:neuron_wise_plots}
 This section provides neuron-wise memorization analysis through various plots.

 \subsection{VAR-d16 Model - Memorization Spread (Neuron Index vs. Block)}
\begin{figure}
 \centering
 \includegraphics[width=0.5\textwidth]{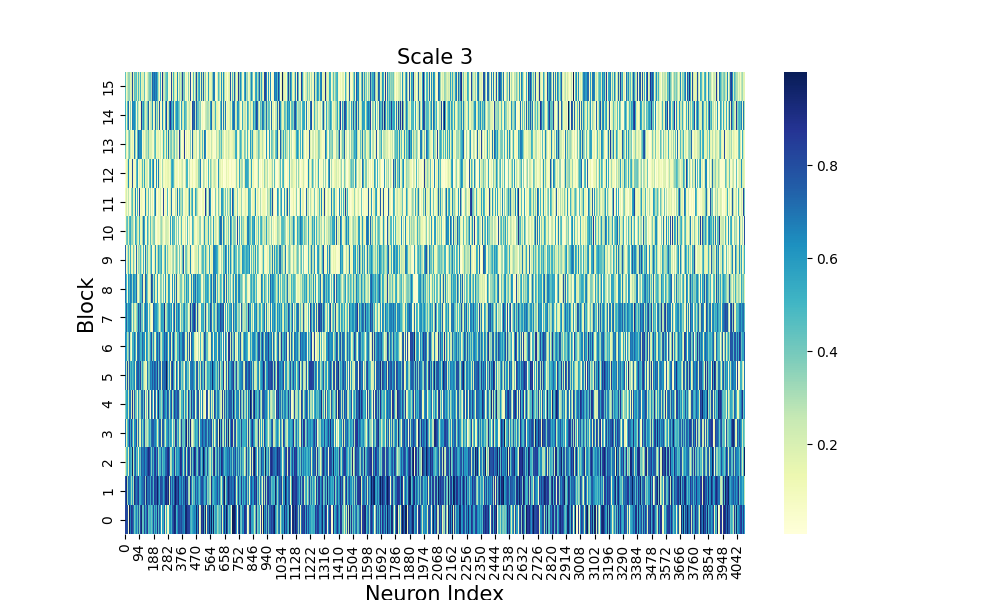} %
 \caption{VAR-d16: Neuron-wise UnitMem scores (fc1 layers) by block - Scale 3.}
 \label{fig:var_d16_neuron_block_s3}
 \end{figure}

\begin{figure}
 \centering
 \includegraphics[width=0.5\textwidth]{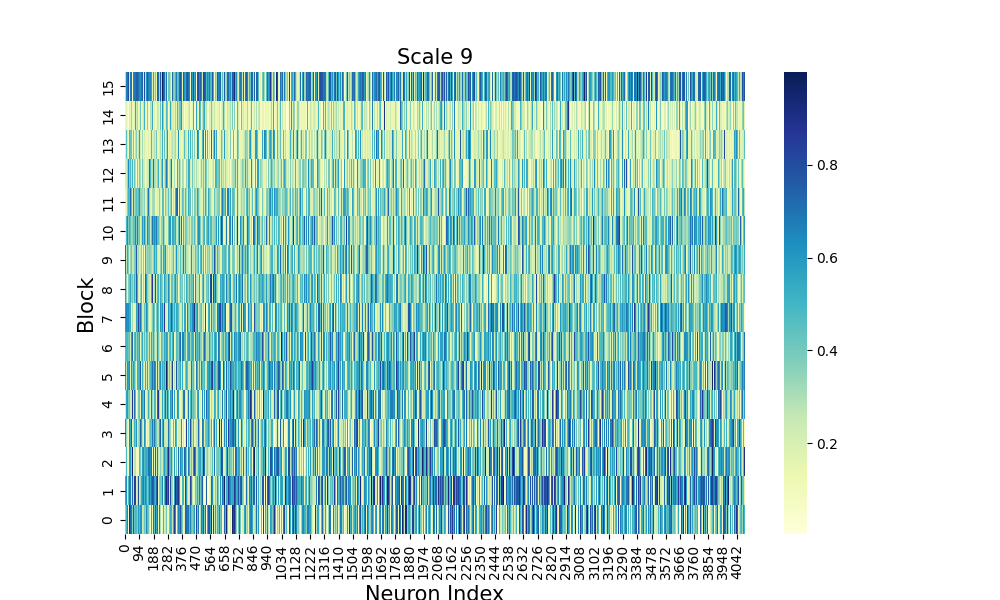} %
 \caption{VAR-d16: Neuron-wise UnitMem scores (fc1 layers) by block - Scale 9.}
 \label{fig:var_d16_neuron_block_s9}
 \end{figure}

 The following Figures \ref{fig:var_d16_neuron_block_s1}\ref{fig:var_d16_neuron_block_s6}\ref{fig:var_d16_neuron_block_s9}\ref{fig:var_d16_neuron_block_s3} for VAR-d16 visualize UnitMem scores for individual neurons (in fc1 layers) across blocks for specific scales. The x-axis represents the neuron index within a block, and the y-axis represents the block index. These plots illustrate how memorization is distributed across neurons and blocks, and how this distribution evolves with increasing generation scale, showing how memorization is diluted and spread as the scale increases.

 \begin{figure}
 \centering
 \includegraphics[width=0.5\textwidth]{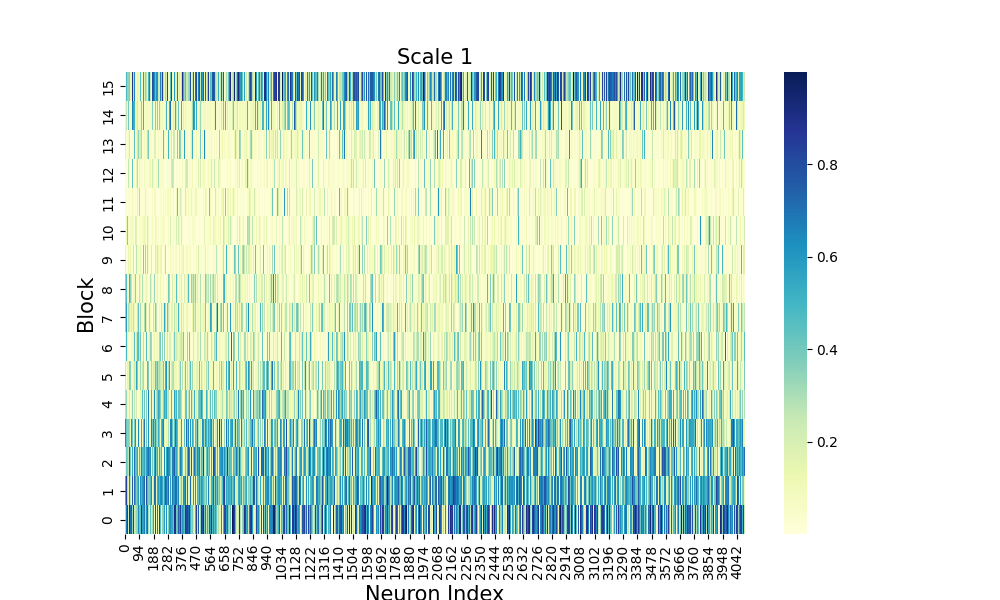} %
 \caption{VAR-d16: Neuron-wise UnitMem scores (fc1 layers) by block - Scale 1. (X-axis: Neuron Index within block, Y-axis: Block Index, Color: UnitMem score). This visualization shows the intensity of memorization for each neuron.}
 \label{fig:var_d16_neuron_block_s1}
 \end{figure}

 \begin{figure}
 \centering
 \includegraphics[width=0.5\textwidth]{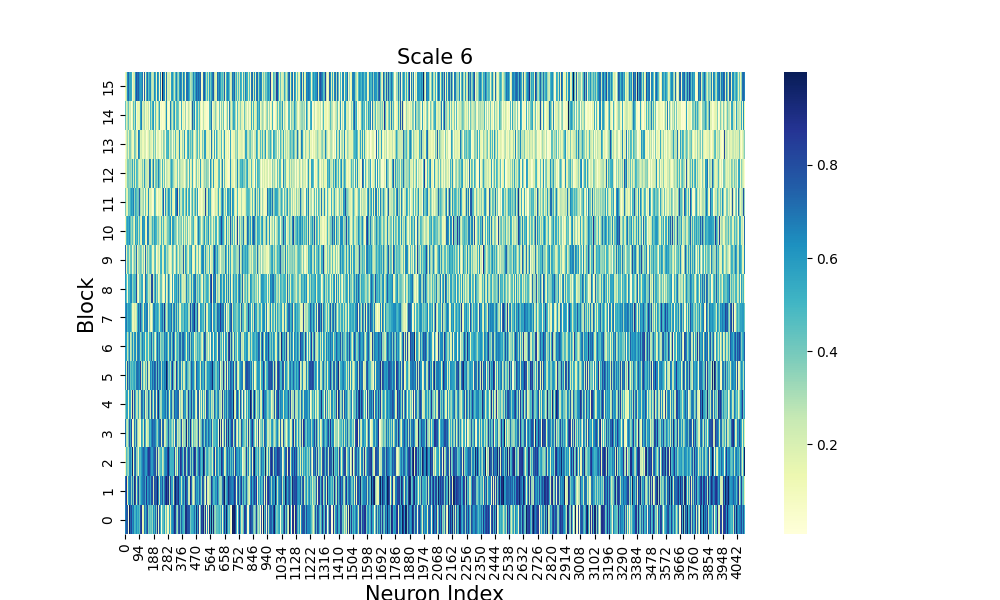} %
 \caption{VAR-d16: Neuron-wise UnitMem scores (fc1 layers) by block - Scale 6.}
 \label{fig:var_d16_neuron_block_s6}
 \end{figure}

 \subsection{RAR Model - UnitMem Histogram}
\begin{figure}
 \centering
 \includegraphics[width=0.5\textwidth]{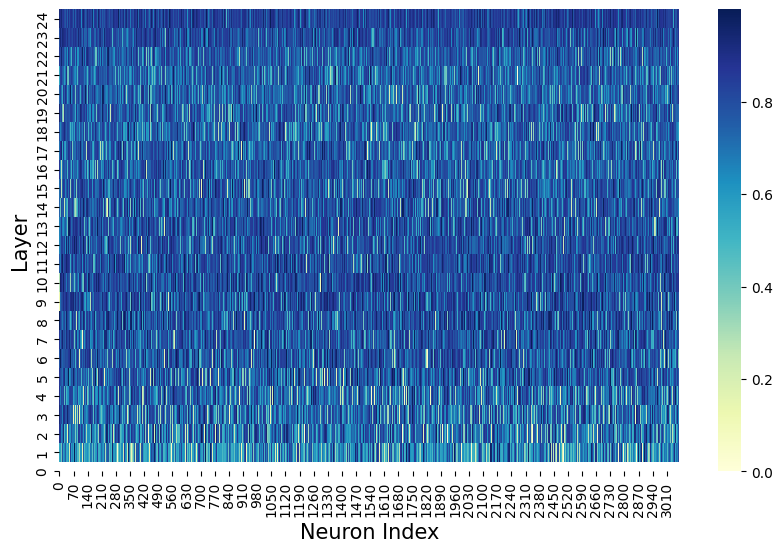} %
 \caption{RAR: Neuron-wise UnitMem distribution in fc1 layers (last token prediction).}
 \label{fig:rar_neuron_hist}
 \end{figure}

 For RAR-Base model, \Cref{fig:rar_neuron_hist} shows the distribution of UnitMem scores across neurons (in fc1 layers) based on the last token prediction.

\end{document}